\documentclass[conference]{IEEEtran}

\usepackage{xcolor}
\definecolor{acceptgray}{gray}{0.55}

\IEEEoverridecommandlockouts
\makeatletter
\IEEEsettopmargin{t}{1in}
\makeatother
\usepackage{cite}
\usepackage{amsmath,amssymb,amsfonts}
\usepackage{graphicx}
\usepackage{textcomp}
\usepackage{tabularx}
\usepackage{booktabs}
\usepackage{multirow}
\usepackage{xcolor}
\usepackage{url}
\usepackage{units}
\usepackage{balance}

\usepackage{tikz}
\usepackage{pgfplots}
\usepackage{pgfmath}

\usepackage{pgfplotstable}
\usetikzlibrary{calc, bending, external, plotmarks, shapes, pgfplots.statistics, pgfplots.colorbrewer, positioning, arrows.meta, matrix, shapes.geometric, decorations.markings, patterns}
\usepgfplotslibrary{fillbetween}
    
\pgfplotsset{scaled y ticks=false}
\pgfplotsset{compat=1.18}

\pgfdeclarelayer{bg}   
\pgfdeclarelayer{fg}   
\pgfsetlayers{bg,main,fg}

\usepackage{array}
\newcolumntype{Y}{>{\raggedright\arraybackslash}X}

\def\BibTeX{{\rm B\kern-.05em{\sc i\kern-.025em b}\kern-.08em
    T\kern-.1667em\lower.7ex\hbox{E}\kern-.125emX}}
\begin{document}

\title{
\vspace{-1.0em}
  {\normalfont
   \fontsize{11pt}{13pt}\selectfont
   \color{acceptgray}
   This paper has been accepted for publication at the\\[-0.1em]
   International Conference on Unmanned Aircraft Systems (ICUAS), 2026.\par
  }
  \vspace{1.2em}
EAAE: Energy-Aware Autonomous Exploration \\ for UAVs in Unknown 3D Environments
}

\author{
Jacob Elskamp$^{1}$,
Moji Shi$^{1}$,
Leonard Bauersfeld$^{2}$,
Davide Scaramuzza$^{2}$,
Marija Popovi\'c$^{1}$

\thanks{$^{1}$J. Elskamp, M. Shi, and M. Popovi\'c are with the Faculty of Aerospace Engineering, Delft University of Technology, Delft, The Netherlands. {\tt\small jacobelskamp@hotmail.com}, {\tt\small \{M.Shi, M.Popovic\}@tudelft.nl}}

\thanks{$^{2}$L. Bauersfeld and D. Scaramuzza are with the Institute for Informatics, University of Zurich, Zurich, Switzerland. Their work was supported by the European Union’s Horizon Europe Research and Innovation Programme under grant agreement No. 101120732 (AUTOASSESS).}}

\maketitle

\begin{abstract}
Battery-powered multirotor unmanned aerial vehicles (UAVs) can rapidly map unknown environments, but mission performance is often limited by energy rather than geometry alone. Standard exploration policies that optimise for coverage or time can therefore waste energy through manoeuvre-heavy trajectories. In this paper, we address energy-aware autonomous 3D exploration for multirotor UAVs in initially unknown environments. We propose Energy-Aware Autonomous Exploration (EAAE), a modular frontier-based framework that makes energy an explicit decision variable during frontier selection. EAAE clusters frontiers into view-consistent regions, plans dynamically feasible candidate trajectories to the most informative clusters, and predicts their execution energy using an offline power estimation loop. The next target is then selected by minimising predicted trajectory energy while preserving exploration progress through a dual-layer planning architecture for safe execution. We evaluate EAAE in a full exploration pipeline with a rotor-speed-based power model across simulated 3D environments of increasing complexity. Compared to representative distance-based and information gain-based frontier baselines, EAAE consistently reduces total energy consumption while maintaining competitive exploration time and comparable map quality, providing a practical drop-in energy-aware layer for frontier exploration.
\end{abstract}


\section{Introduction}
Autonomous exploration is a key capability for unmanned aerial vehicles (UAVs) operating in unknown or partially known environments. It is central to applications such as search and rescue, infrastructure inspection, and remote surveillance, where manual teleoperation is inefficient, cognitively demanding, or unsafe \cite{SAR_uavs,Industrial_UAVs,dronesPlanetary}. In these settings, small multirotor UAVs offer high mobility and access, but their operation is fundamentally constrained by limited onboard battery capacity. Consequently, exploration systems should not only maximise environment coverage or reduce mission duration, but also use the available energy efficiently.

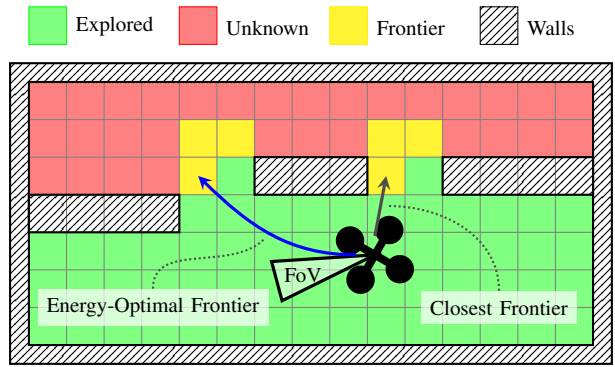
\begin{figure}[ht!]
    \centering
    \begin{tikzpicture}[>=stealth]
        \tikzset{every node/.style={font=\footnotesize}}
        \tikzstyle{gridlines} = [draw, white!50!black]
        \def\w{7.5}
        \def\h{3.5}

        \begin{pgfonlayer}{bg}
            \draw[thick, pattern=north east lines] (-0.25,-0.25) rectangle (\w+0.25,\h+0.25); 
            \draw[fill=white] (0, 0) rectangle (\w, \h);
        \end{pgfonlayer}
        
        \foreach \x in {0,0.5,...,\w} {
            \draw [gridlines] (\x,0) -- ++ (0, \h);
        }
        \foreach \y in {0,0.5,...,\h} {
            \draw [gridlines] (0,\y) -- ++ (\w, 0);
        }
        \draw[thick] (0, 0) rectangle (\w, \h);

        \draw [thick, pattern=north east lines] (0, 1.5) rectangle ++ (2, 0.5);
        \draw [thick, pattern=north east lines] (3, 2.0) rectangle ++ (1.5, 0.5);
        \draw [thick, pattern=north east lines] (5.5, 2.0) rectangle ++ (2, 0.5);

        \begin{pgfonlayer}{bg}
            \fill [green, fill opacity=0.5] (0,0) rectangle (\w,1.5);
            \fill [green, fill opacity=0.5] (2,1.5) rectangle (\w,2.0);
            \fill [green, fill opacity=0.5] (2.5,2.0) rectangle ++ (0.5, 0.5);
            \fill [green, fill opacity=0.5] (5,2.0) rectangle ++ (0.5, 0.5);
        \end{pgfonlayer}
        
        \begin{pgfonlayer}{bg}
            \fill [red, fill opacity=0.5] (0,2) rectangle ++ (2,1.5);
            \fill [red, fill opacity=0.5] (3,2.5) rectangle ++ (1.5,1.0);
            \fill [red, fill opacity=0.5] (5.5,2.5) rectangle ++ (2,1.0);
            \fill [red, fill opacity=0.5] (2.0,3.0) rectangle ++ (1,0.5);
            \fill [red, fill opacity=0.5] (4.5,3.0) rectangle ++ (1,0.5);
        \end{pgfonlayer}
        
        \begin{pgfonlayer}{bg}
            \fill [yellow, fill opacity=0.8] (2.0,2.5) rectangle ++ (1,0.5);
            \fill [yellow, fill opacity=0.8] (4.5,2.5) rectangle ++ (1,0.5);
            \fill [yellow, fill opacity=0.8] (2.0,2.0) rectangle ++ (0.5,0.5);
            \fill [yellow, fill opacity=0.8] (4.5,2.0) rectangle ++ (0.5,0.5);
        \end{pgfonlayer}

        \begin{scope}[xshift=4.6cm, yshift=1.2cm, scale=0.27, rotate=15]
            \fill [black] (1,1) circle (0.7);
            \fill [black] (-1,1) circle (0.7);
            \fill [black] (1,-1) circle (0.7);
            \fill [black] (-1,-1) circle (0.7);
            \draw [line width=3pt] (1, 1) -- (-1, -1);
            \draw [line width=3pt] (-1, 1) -- (1, -1);
            \filldraw [very thick, fill=white, fill opacity=0.5] (0,0) -- (-5, -1) -- ++ (0, 2) -- cycle;
            \node [anchor = west] at ([yshift=9pt]-5, 0) {FoV};
            \coordinate (drone) at (0,0);
        \end{scope}

        \draw [very thick, blue, ->] ([xshift=-0.25cm]drone) to [out=180, in = -45] coordinate [midway] (line1) (2.25,2.25);
        \draw [very thick, black!70!white, ->] ([yshift=0.25cm]drone) -- coordinate [midway] (line2) (4.75, 2.25);

        \filldraw [draw=gray, green, fill opacity=0.5] (0, \h+0.5) rectangle ++(0.5, 0.5);
        \filldraw [draw=gray, red, fill opacity=0.5] (2, \h+0.5) rectangle ++(0.5, 0.5);
        \filldraw [draw=gray, yellow, fill opacity=0.8] (4, \h+0.5) rectangle ++(0.5, 0.5);
        \draw [pattern=north east lines] (6, \h+0.5) rectangle ++(0.5, 0.5);
        \node [anchor = south west] at (0.5, \h + 0.5) {Explored};
        \node [anchor = south west] at (2.5, \h + 0.5) {Unknown};
        \node [anchor = south west] at (4.5, \h + 0.5) {Frontier};
        \node [anchor = south west] at (6.5, \h + 0.5) {Walls};

        \node (fclose) [fill = white, fill opacity = 0.7, text opacity = 1.0] at (6.25, 0.5) {Closest Frontier};
        \node (fopt) [fill = white, fill opacity = 0.7, text opacity = 1.0] at (1.65, 0.5) {Energy-Optimal Frontier};
        \draw [thick, densely dotted, black!70!white]  ([yshift = 3pt]fclose) to [out = 90, in=0] ([xshift=3pt]line2);
        \draw [thick, densely dotted, black!70!white]  ([yshift = 3pt]fopt) to [out = 90, in=-135] ([xshift=-2pt, yshift=-2pt]line1);
    \end{tikzpicture}
    \caption{We introduce EAAE, an energy-aware decision layer for frontier-based exploration using UAVs. By integrating trajectory-level energy prediction into frontier selection, EAAE enables more energy-efficient exploration. Here, EAAE selects an energy-optimal frontier (blue) instead of the geometrically closest one (grey) under the same field of view (FoV) constraints.}
    \label{fig:fig1}
    \vspace*{-12pt}
\end{figure}

In this paper, we address the problem of \emph{energy-aware autonomous exploration for multirotor UAVs in unknown 3D environments}. Standard exploration pipelines typically prioritise the next target based on  geometric heuristics like nearest-frontier proximity~\cite{yamauchi1998frontier} or utility-based ranking based on predicted map gain and path cost~\cite{julia2012comparison,Bircher2016}. While these criteria support efficient environment coverage, they do not explicitly model the energy cost of the resulting flight trajectory. This is a critical limitation for multirotor platforms, whose energy consumption depends not only on the distance travelled, but also on maneuvering behavior, accelerations, heading changes, and altitude transitions~\cite{bauersfeld2022range}. As a result, trajectories with similar geometric length may induce substantially different energy expenditures, which directly affects achievable exploration progress under endurance constraints as illustrated in Fig.~\ref{fig:fig1}.

To address this, a common strategy is to augment frontier-based exploration with execution-aware criteria. Recent methods employ a hierarchical planning framework to reach selected geometric frontiers~\cite{Zhou2021,Zhao2024,hou2024laea}, yet they typically treat energy only indirectly through time or distance proxies. In contrast, our approach explicitly estimates the expected energy expenditure of candidate trajectories prior to selection.

The main contribution of this paper is \emph{Energy-Aware Autonomous Exploration (EAAE)}, a modular frontier-based exploration framework that integrates trajectory-level energy prediction directly into UAV frontier selection for autonomous 3D exploration. Rather than redesigning the exploration stack, EAAE introduces energy awareness as an explicit decision layer on top of established mapping, frontier extraction, clustering, and trajectory-planning components. This makes the approach practically relevant and easy to integrate into existing frontier-based systems, while enabling principled target selection based on predicted execution cost. We implement EAAE in a complete exploration pipeline and evaluate it in simulated environments against representative baselines that do not account for energy efficiency during decision-making.

This paper makes two contributions:
\begin{enumerate}
    \item We propose EAAE, an energy-aware frontier exploration framework for unknown 3D environments that selects among frontier clusters using trajectory-level energy prediction for multirotor UAVs.
    \item We evaluate a planning architecture leveraging EAAE that couples global kinodynamic trajectory generation with local reactive planning for safe execution, demonstrating improved energy efficiency over standard non-energy-aware baselines.
\end{enumerate}

\section{Related Work}

Exploration is a core capability of autonomous unmanned aerial vehicles (UAVs). It refers to the process of uncovering unknown space in an initially unknown environment by repeatedly selecting informative targets and navigating to them while updating a map. While most existing methods prioritise time, distance, or information gain, they often neglect the actual energy cost of the flight dynamics. This section reviews current exploration frameworks and highlights the need for energy-aware target selection to improve operational efficiency.

\textbf{Frontier-based exploration.} Frontier exploration was formalised by Yamauchi~\cite{yamauchi1998frontier} and remains a dominant strategy for unknown-map exploration due to its simplicity and strong coverage performance~\cite{julia2012comparison}. The key concept is to target frontiers, defined as the boundaries between known free space and unmapped regions. Many frontier methods use greedy next-target selection based on geometric heuristics such as distance, frontier size, or simple turning penalties~\cite{topiwala2018frontier,gao2018improved}. Several extensions improve practical performance: velocity-aware frontier selection to enable faster flight~\cite{Cieslewski2017}, hierarchical frontier structures and incremental replanning (FUEL)~\cite{Zhou2021}, and adaptive global sequencing and yaw strategies to reduce back-and-forth behaviour (FAEP)~\cite{Zhao2024}. Non-greedy formulations based on route optimisation have also been explored~\cite{Meng2017TSP}. While these methods substantially improve exploration efficiency, their target-ranking criteria typically remain geometric (or time-proxy) and therefore do not explicitly account for trajectory-dependent energy expenditure that can vary significantly even for similar path lengths on multirotor platforms~\cite{bauersfeld2022range}.

\textbf{Sampling, NBV, and hybrid exploration.} Next-best-view (NBV) methods optimise viewpoint quality, such as information gain, while being subject to a travel cost term, e.g. trajectory time or distance~\cite{Bircher2016, HectorH2002, popovic2024learning}. While advantageous in cluttered environments where visibility is critical, NBV can become computationally intensive in large-scale maps~\cite{caiza2024autonomous}. Hybrid approaches attempt to balance this by combining frontier-based global coverage with local sampling~\cite{selin2019efficient}. Despite these sophisticated selection strategies, the underlying cost functions typically assume a constant power-to-speed relationship. They consequently fail to account for the high-power transients associated with the aggressive accelerations and frequent heading changes inherent to multirotor flight in complex 3D spaces~\cite{bauersfeld2022range}.

\textbf{Energy-aware exploration.} While energy-aware planning for UAVs is a rapidly emerging field, it remains less established than coverage- or information-driven strategies. Prior work has introduced empirical phase-based cost models~\cite{CarmeloDiFranco2015}, distance-penalty approximations~\cite{wang2024efficient, rappaport2016energy}, and speed-efficiency insights~\cite{Cieslewski2017, patel2023towards}. These efforts demonstrate that energy consumption cannot be reliably inferred from geometric distance alone~\cite{bauersfeld2022range}. However, in exploration settings, explicit target selection based on the predicted energy of full candidate trajectories while maintaining online reactivity remains a significant challenge. Our proposed approach, EAAE, addresses this gap by integrating trajectory-level energy prediction directly into the frontier selection loop, enabling energy-aware decisions without requiring a redesign of the underlying exploration stack. This allows the UAV to trade off information gain against the high power costs of 3D maneuvers, thereby extending mission endurance in energy-constrained exploration scenarios.

\section{Our Approach}
\begin{figure*}[t!]
    \centering
    \begin{tikzpicture}[>=stealth]
        \tikzset{every node/.style={font=\footnotesize}}
        \tikzstyle{innerbox} = [draw, fill=white, text width = 2.3cm, minimum width = 2.6cm, minimum height=0.85cm, align=center, outer sep = 4pt]
        \def\dx{5pt}
        
        \node [innerbox] (point_cloud) {Point Cloud};
        \node [right = 0mm of point_cloud, innerbox] (occupancy_map) {Occupancy Map};
        \node [anchor = south west] (perc_map_lbl) at ([xshift=-3mm]point_cloud.north west) {\textbf{Perception and Mapping}};
        \begin{pgfonlayer}{bg}
            \filldraw [draw, fill = black!5!white] (perc_map_lbl.north west) rectangle (occupancy_map.south east);
        \end{pgfonlayer}

        \node [right = 10mm of occupancy_map, innerbox] (frontier) {Frontier detection and clustering};
        \node [right = 5mm of frontier, innerbox] (energy_module) {Energy-aware module};
        \node [right = 5mm of energy_module, innerbox] (target_selection) {Final target selection};
        \node [anchor = south west] (expl_lbl) at ([xshift=-3mm]frontier.north west) {\textbf{Exploration Algorithm (Main Contribution)}};
        \begin{pgfonlayer}{bg}
            \filldraw [draw = blue, thick, fill = black!5!white] (expl_lbl.north west) rectangle ([xshift=0.5cm, yshift=-0.4cm]target_selection.south east);
        \end{pgfonlayer}

        \node [below = 5mm of frontier, innerbox, draw=none] (dummy_node) {};
        \node [right = 5mm of dummy_node, innerbox] (global_planner) {Global Planner \& Energy Estimator};
        \node [right = 5mm of global_planner, innerbox] (local_planner) {Reactive local planner};

        \node [anchor = north west] (planning_lbl) at ([xshift=-3mm]dummy_node.north west) {\textbf{Planning module}};
        \begin{pgfonlayer}{bg}
            \draw [draw, dashed] (planning_lbl.north west) rectangle ([xshift=0.5cm, yshift=-12pt]local_planner.south east);
        \end{pgfonlayer}
      
        \node [below = 7mm of point_cloud, innerbox] (uav) {UAV};
        \node [right = 0mm of uav, innerbox] (agiros) {Agiros pilot};
        \node [anchor = south west] (control_lbl) at ([xshift=-3mm]uav.north west) {\textbf{Control and simulation}};
        \begin{pgfonlayer}{bg}
            \filldraw [draw, fill = black!5!white] (control_lbl.north west) rectangle (agiros.south east);
        \end{pgfonlayer}

        \draw [thick, ->] (occupancy_map) -- ([xshift=4pt]frontier.west);
        \path (uav) -- coordinate [midway] (midpoint_ctrl)  (agiros);
        \draw [thick, ->] (control_lbl.north -| midpoint_ctrl) -- (midpoint_ctrl |- occupancy_map.south); 
        \draw [thick, ->] ([yshift=4pt]local_planner.south) |- node [near end, below=-0.5mm] {\emph{Final trajectory}} (local_planner.south -| planning_lbl.south) |- ([xshift=-4pt]agiros.east);
        \draw [->] ([xshift=-0.25cm,yshift=-4pt]global_planner.north) -- node [pos=1, anchor = north east] {\emph{Candiates}} ([xshift=-0.25cm,yshift=4pt]energy_module.south);
        \draw [<-] ([xshift=0.25cm,yshift=-4pt]global_planner.north) -- node [pos=1, anchor = north west] {\emph{Energy consumption}} ([xshift=0.25cm,yshift=4pt]energy_module.south);
        \draw [->] ([yshift=-4pt]local_planner.north) -- node [pos=1, anchor = north west] {\emph{Goal Waypoint}} ([yshift=4pt]target_selection.south);
    \end{tikzpicture}
    \caption{We introduce EAAE, a frontier-based exploration framework that enables energy-aware goal selection by predicting the energy of candidate trajectories. Our pipeline fuses UAV depth data into an OctoMap (Sec. III-A), clusters frontiers into goals (Sec. III-B), estimates energy offline (Sec. III-C), and executes the chosen goal with reactive local planning (Sec. III-D). By integrating energy awareness into planning, EAAE favours informative frontiers that are cheaper to reach.}
    \label{fig:system_overview}
\end{figure*}
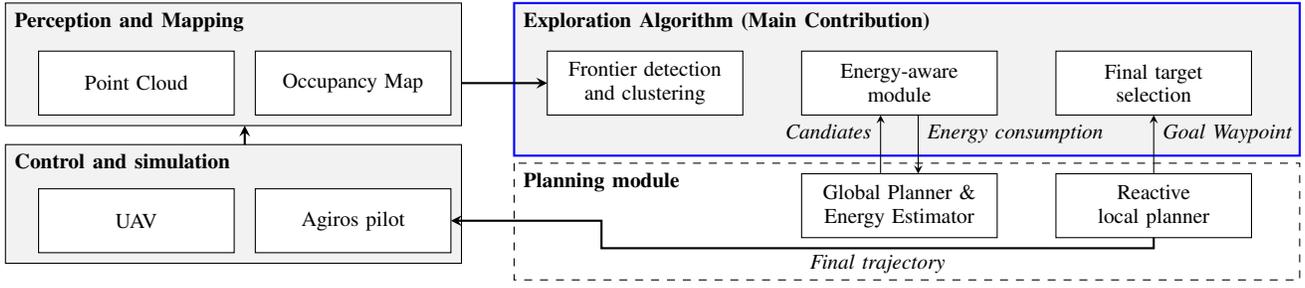

We propose \emph{Energy-Aware Autonomous Exploration (EAAE)}, a modular extension to frontier-based exploration that makes energy an explicit decision variable during target selection for multirotor UAVs in unknown 3D environments. Fig. \ref{fig:system_overview} shows an overview of our approach. EAAE follows an iterative loop: (i) we update an octomap-based occupancy representation from onboard depth sensing, (ii) we extract and cluster frontiers into view-consistent candidate regions, (iii) we generate dynamically feasible global trajectories to a reduced set of high-utility candidates, (iv) we estimate the energy required to execute each candidate trajectory via offline power evaluation, and (v) we execute the selected goal using a local reactive planner for safety.

A key design choice is to separate decision-time energy reasoning from execution-time reactivity. Rather than approximating energy using distance or time, EAAE predicts the trajectory-level execution cost for each candidate using an offline simulation loop, and selects the next goal by minimising predicted energy among feasible high-information candidates. At runtime, the UAV executes the selected goal with a dual-layer planning architecture: a global kinodynamic planner provides dynamically feasible candidate trajectories for evaluation, while a local reactive planner handles obstacle avoidance and map changes during flight. The remainder of this section details the different components of our pipeline.


\subsection{UAV Perception and Environment Mapping}
We maintain a probabilistic 3D occupancy map of the environment using an OctoMap-style voxel grid~\cite{OctoMap_hornung2013octomap}. The workspace is discretised into cubic voxels of resolution $r$, which sets the trade-off between geometric detail and memory/computation. The UAV observes the scene with a forward-facing depth camera, producing point clouds that are integrated online to update voxel occupancy probabilities. Each voxel is classified as \emph{free}, \emph{occupied}, or \emph{unknown}, yielding the partition:
\begin{equation}
V_{\text{total}} \equiv V_{\text{free}} \cup V_{\text{occupied}} \cup V_{\text{unknown}}.
\end{equation}
During exploration, $V_{\text{unknown}}$ monotonically decreases. This representation supports efficient spatial queries for both motion planning and frontier extraction, and it provides a consistent basis for tracking exploration progress. We consider exploration complete when the remaining unknown volume is limited to residual regions that are not observable due to sensor visibility constraints, e.g. occlusions.

\subsection{Frontier Detection and Clustering}

Based on the Octomap, a frontier voxel $f\in\mathcal{F}$ is defined as a free voxel that has at least one neighboring unknown voxel~\cite{yamauchi1998frontier}. Frontier voxels are therefore reachable through known free space while bordering unexplored regions, making them natural candidates for informative exploration goals. In EAAE, we iterate over map voxels and use OctoMap utilities to test whether a voxel is free and whether at least one of its neighbours is unknown. Voxels satisfying both conditions are added to the frontier set.

At practical map resolutions, the complete frontier set can become large, making direct evaluation of all frontier voxels computationally inefficient. To reduce the number of candidate exploration targets while preserving the information content of the frontier structure, EAAE applies divisive $k$-means clustering to the detected frontier set. Unlike standard $k$-means, this top-down clustering strategy avoids specifying the final number of clusters a priori, which is advantageous in exploration problems where the environment size and the number of frontiers are unknown. Starting from the full frontier set, the method recursively splits each set into two sub-clusters based on Euclidean distance until all clusters satisfy a maximum size requirement.

\begin{figure}[ht!]
    \centering
    \begin{tikzpicture}[>=stealth]
       \tikzset{every node/.style={font=\footnotesize}}
       \def\l{5}
       \def\w{1.7}
       \def\r{2pt}
       \coordinate (c0) at (0,0);
       \coordinate (c1) at (\l, 0);
       \coordinate (l1) at (\l, -\w);
       \coordinate (u1) at (\l, \w);

        \begin{scope}[xshift = \l cm]
            \pgfmathsetseed{43}
            \foreach \i in {1,...,20} {
                \pgfmathsetmacro{\x}{2 * (rnd-0.5)}
                \pgfmathsetmacro{\y}{2 * (rnd-0.5)}
                \fill [yellow] (\x,\y) circle (\r);
            }
            \node (ctr) at (0.1, 0.2) [circle, draw=none, fill=red, inner sep = 0pt, outer sep = 0pt, minimum width = 4pt] {} ;
            \node (max) at (0.85, 0.9) [circle, draw=none, fill=yellow, inner sep = 0pt, outer sep = 0pt, minimum width = 4pt] {} ;
            \node [anchor=east] (ctr_lbl) at ([xshift=-0.5cm, yshift=0.5cm]ctr) {cluster centroid};
            \draw [->] (ctr_lbl) -- (ctr);
            \fill [yellow] (max) circle (\r);
            \draw [dashed, <->] (ctr) -- node [midway, below right] {$r_\text{max}$} (max);
        \end{scope}
        \draw [dashed] (c0) -- node [midway, above] {$d_\text{max}$} (c1);
        \draw [thick, densely dotted] (c0) -- (l1);
        \draw [thick, densely dotted] (c0) -- (u1);
        \draw [dashed, <->] (l1) -- node [midway, right] {$\tan(\text{FoV}/2) d_\text{max}$} (c1);
        \node [anchor = east] at (c0) {UAV};
        \node at (\l, 0.75*\w) {frontier points};

        \def\rdraw{2}
        \def\angle{18.77}
        \draw [<->] (c0) ++(-\angle:\rdraw) arc[start angle=-\angle, end angle=\angle, radius=\rdraw] node [near start, right] {FoV};
        
    \end{tikzpicture}
    \caption{Divisive $k$-means cutoff geometry used to enforce view-consistent frontier clusters.}
    \label{fig:cluster_max_distance}
\end{figure}
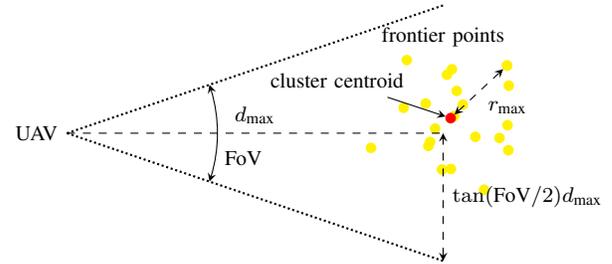

We define this stopping criterion through a field of view (FoV) constraint:
\begin{equation}
r_{\max} \le \tan\left(\frac{\mathrm{FoV}_{\mathrm{hor}}}{2}\right)d_{\max},
\label{eq:cluster_cutoff}
\end{equation}
where $r_{\max}$ denotes the maximum Euclidean distance between any frontier point and the centroid of its cluster, $\mathrm{FoV}_{\mathrm{hor}}$ is the horizontal FoV of the depth camera, and $d_{\max}$ is the maximum camera sensing range. This constraint ensures that the full spatial extent of a cluster can be observed from a single viewpoint, which reduces redundant repositioning and supports efficient sensing. The corresponding geometric interpretation is illustrated in Fig.~\ref{fig:cluster_max_distance}.

The clustering procedure has two practical benefits. First, the divisive strategy removes the need to predefine the number of clusters. Second, it reduces the number of candidate waypoints while retaining exploration-relevant information, since the representation preserves both the cluster centroid and the cluster size, i.e. number of frontier voxels in the cluster.

After clustering, we filter candidate clusters for feasibility. Specifically, we check reachability around each cluster centroid by examining neighbouring voxels and rejecting candidates whose surrounding space contains known occupied voxels. This avoids selecting goals that are infeasible or likely to produce collision-prone trajectories. To support the subsequent global planning module, we then sample multiple candidate viewpoints around each feasible cluster centroid and retain only those located in free space. For every sampled viewpoint, we compute a target yaw angle such that the UAV is oriented toward the corresponding cluster centroid. As a result, when the UAV reaches the selected viewpoint, its onboard sensor is aligned to observe the associated frontier region.

\begin{table*}[t!]
\caption{Energy-estimation strategy comparison relative to the proposed offline execution approach in EAAE.}
\label{tab:energy_methods_comparison_singlecol}
\centering
\small
\begin{tabularx}{1\linewidth}{p{3.0cm}|Yp{6.5cm}} 
\toprule
\textbf{Method} & \textbf{Advantages} & \textbf{Disadvantages} \\
\midrule
Empirical flight-state-based estimation \cite{CarmeloDiFranco2015}
& Fast; simple to implement; does not require a full dynamically feasible trajectory
& Requires empirical data and trajectory segmentation; platform-dependent; limited accuracy \\
\midrule
Velocity-based estimation \cite{Cieslewski2017}
& Relatively lightweight; easy to implement; more generalizable than empirical fitting
& Uses indirect assumptions on achievable flight speed; less suitable in cluttered environments where velocity profiles are constrained by obstacles \\
\midrule
Distance-based estimation \cite{wang2024efficient}
& Very simple; low computational cost; broadly applicable
& Ignores UAV dynamics; assumes energy scales mainly with path length; cannot distinguish aggressive from smooth maneuvers \\
\midrule
Offline execution (this work)
& Evaluates actual planned trajectory; accounts for UAV dynamics and controller response; avoids approximation
& Higher computational cost; requires offline trajectory generation for all candidates \\
\bottomrule
\end{tabularx}
\end{table*}

Because each remaining candidate requires trajectory generation for subsequent energy evaluation, we perform an additional candidate reduction step to limit the computational burden of the global planner. We approximate the exploration utility of each cluster by its information gain:
\begin{equation}
IG(c_i)=\mathrm{COUNT}(f\in c_i),
\end{equation}
where $\mathrm{COUNT}(f\in c_i)$ is the number of frontier voxels contained in frontier cluster $c_i$. This simple proxy favours larger frontier regions expected to reveal more unknown space, while discarding smaller clusters with limited exploration value. Finally, only feasible clusters with the highest information gain are forwarded to the energy-aware planning stage.

\begin{figure}[ht!]
\centering
    \begin{tikzpicture}
        \tikzset{every node/.style={font=\footnotesize}}
        \def\l{0.45\linewidth}
        \tikzstyle{img} = [inner sep=0pt, outer sep=0pt]
        \tikzstyle{capt} = [
            text width = 3cm,
            align=center,
            inner sep = 0pt,
            outer sep=0pt,
            anchor=north,
            yshift=-3pt
        ]
        
        \node[img] (a)
            {\includegraphics[width=\l]{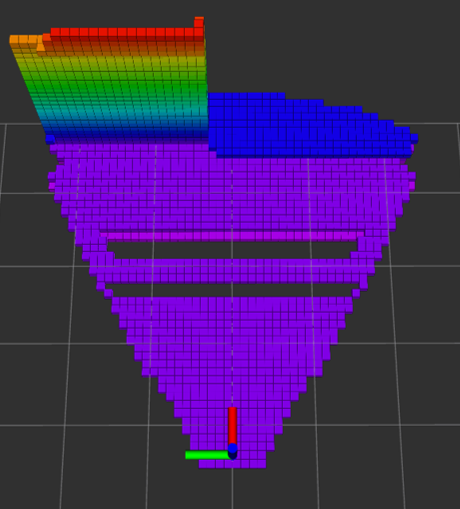}};
        
        \node[img, right=6mm of a] (b)
            {\includegraphics[width=\l]{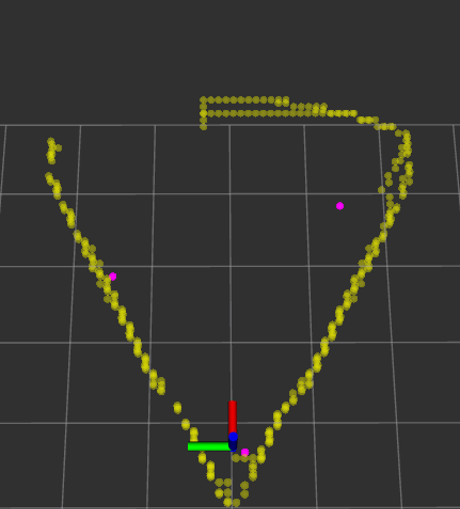}};
    \end{tikzpicture}
    \caption{Frontier extraction and clustering from the map representation. Left: OctoMap occupancy  visualisation. Right: frontier detection and clustering, where yellow voxels denote frontier voxels and pink points denote frontier cluster centroids.}
    \label{fig:octomap_cluster}
\end{figure}

The overall process of map construction, frontier extraction, and frontier clustering is illustrated in Fig.~\ref{fig:octomap_cluster}. The first image shows the OctoMap-based 3D occupancy representation, while the second image visualises the detected frontier voxels and the resulting frontier clusters.

\subsection{Energy-awareness Module}
\label{subsec:energy_aware_module_approach}

After frontier candidates have been detected, clustered, and filtered, the exploration cycle enters the energy-awareness stage. EAAE adopts a two-layer planning architecture consisting of a global planner and a local planner. The global planner is used to generate full candidate trajectories for target evaluation, while the local planner is used during execution for reactive collision avoidance. Specifically, the global planner employs risk-aware kino-dynamic planning to compute dynamically feasible trajectories toward candidate frontier clusters~\cite{Moji_planner}, whereas the local planner uses EGO-planner to adapt online to local obstacles and environmental changes during flight~\cite{zhou2020ego}.

For each remaining candidate cluster, a global trajectory is generated using the offline planner, which computes a complete trajectory from the current UAV state to the candidate target before execution. The resulting candidate trajectories are then evaluated using the \textit{Agilicious} framework~\cite{foehn2022agilicious} to estimate the energy required to reach each candidate.

Offline planning differs from online planning in that the full trajectory is computed prior to execution, without feedback or replanning during flight. In contrast, online planning generates motion incrementally based on sensor measurements and environmental changes, as done by the local reactive planner in EAAE. The key advantage of offline planning in the present context is that it enables trajectory-level reasoning before execution. Since the full state and input evolution are available, candidate trajectories can be compared not only geometrically but also in terms of their expected energy demand. As discussed by Shiller~\cite{shiller2015offline}, offline planners can explicitly account for system dynamics and actuator limits through numerical optimisation, making them well suited for evaluating dynamically feasible motion plans.

\begin{figure*}[t!]
\centering
\begin{tikzpicture}
\tikzset{every node/.style={font=\footnotesize}}
\def\l{0.185\linewidth}
\tikzstyle{img} = [inner sep=0pt, outer sep=0pt]
\tikzstyle{capt} = [text width = 3cm, align=center, inner sep = 0pt, outer sep=0pt, anchor=north, yshift=-3pt]

\node[img] (a)
    {\includegraphics[width=\l]{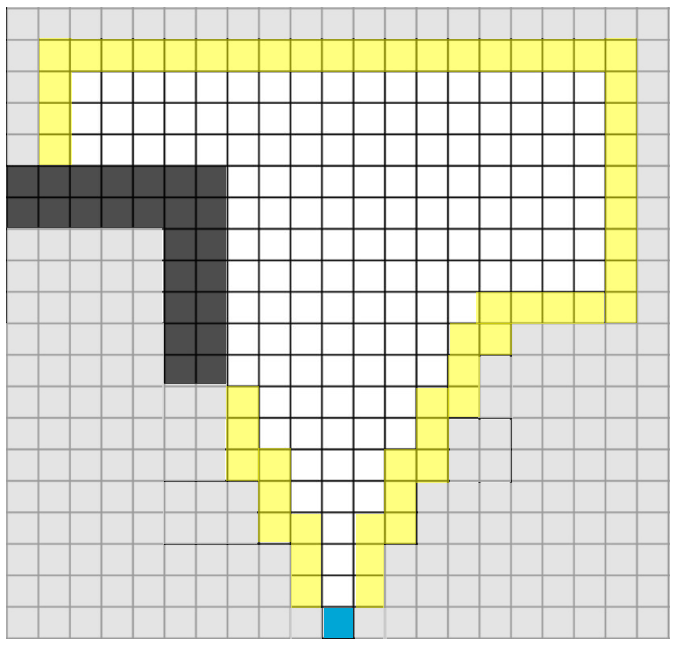}};
\node[capt] at (a.south) {(a) Detect frontiers};

\node[img, right=3mm of a] (b)
    {\includegraphics[width=\l]{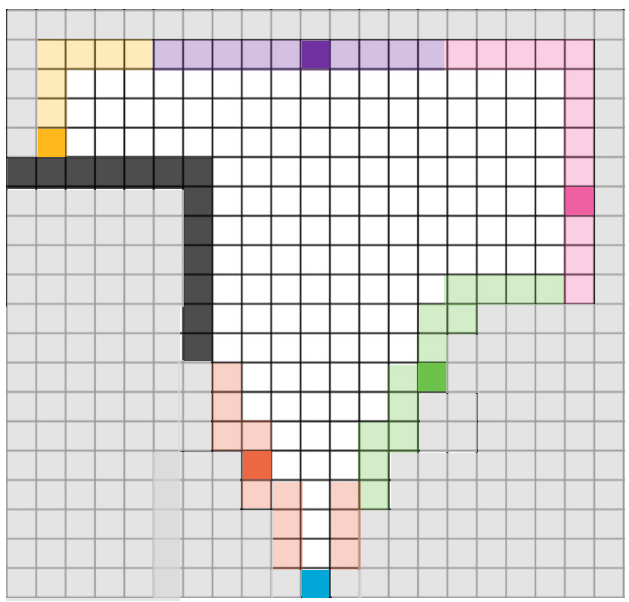}};
\node[capt] at (b.south) {(b) Cluster};

\node[img, right=3mm of b] (c)
    {\includegraphics[width=\l]{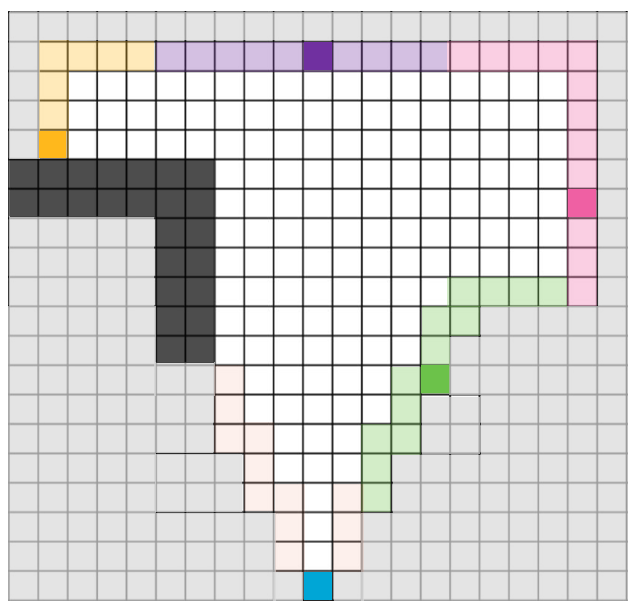}};
\node[capt] at (c.south) {(c) Filter};

\node[img, right=3mm of c] (d)
    {\includegraphics[width=\l]{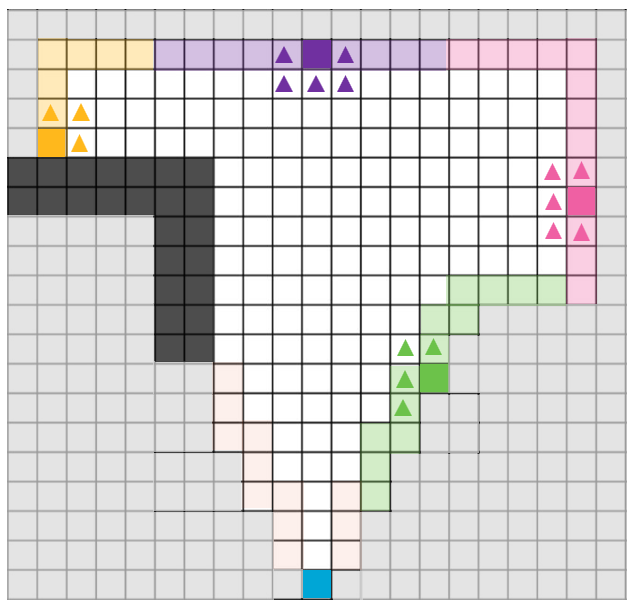}};
\node[capt] at (d.south) {(d) Sample viewpoints};

\node[img, right=3mm of d] (e)
    {\includegraphics[width=\l]{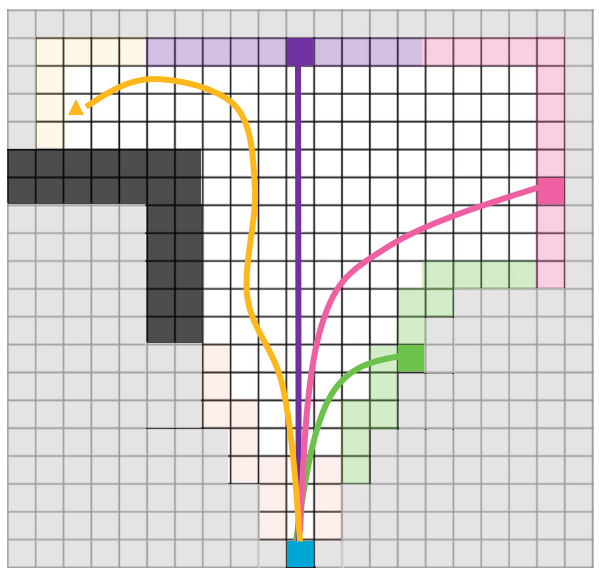}};
\node[capt] at (e.south) {(e) Offline trajectories};

\end{tikzpicture}

\caption{Frontier exploration cycle in EAAE. We: (a) detect frontier voxels from the occupancy map, (b) cluster frontiers via divisive $k$-means into view-consistent regions, (c) filter clusters for feasibility, (d) sample collision-free viewpoints with yaw aligned toward each cluster, and (e) generate candidate global trajectories for offline energy evaluation.}
\label{fig:exploration_cycle}

\end{figure*}
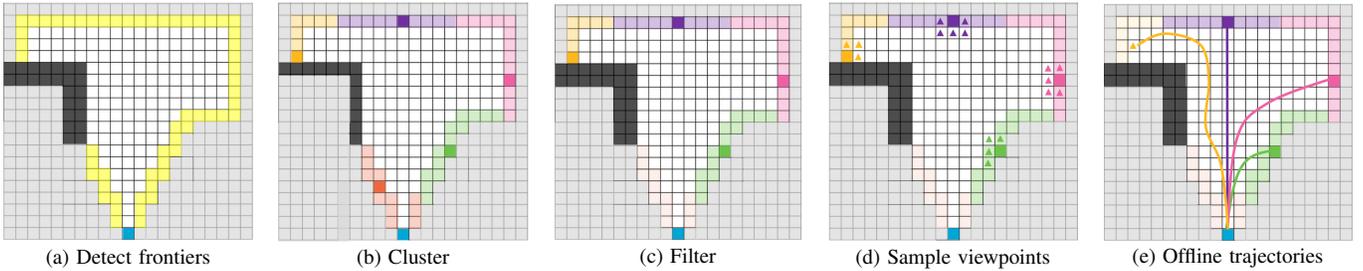

To motivate this design choice, Tab.~\ref{tab:energy_methods_comparison_singlecol} compares our proposed offline execution strategy with common alternatives for energy-aware UAV planning. Existing methods typically estimate energy indirectly, e.g. from empirical flight states, velocity-related metrics, or geometric distance. While computationally efficient, these approaches rely on simplified assumptions that may not hold in cluttered exploration scenarios.

The comparison highlights the trade-off between computational efficiency and estimation fidelity. Empirical flight-state-based methods~\cite{CarmeloDiFranco2015} are simple and fast, but their dependence on platform-specific data reduces transferability and limits accuracy. Velocity-based methods~\cite{Cieslewski2017} provide a lightweight alternative, yet they assume that energy can be inferred from favourable speed profiles, which may not hold when the UAV must manoeuvre in constrained spaces. Distance-based methods~\cite{wang2024efficient} are even simpler, but they reduce the problem to a purely geometric one and fail to capture the effects of acceleration, deceleration, and turning. For this reason, we adopt offline trajectory execution for energy estimation. Rather than relying on a proxy metric, we evaluate the candidate trajectory produced by the planner, making the estimate sensitive to the full motion profile. The offline trajectory generation is visualised in Fig. \ref{fig:exploration_cycle}.

After generating candidate trajectories, we pass them to the energy estimation module. Each trajectory is executed offline in the \textit{Agisim} physics simulator of the \textit{Agilicious} framework~\cite{foehn2022agilicious}. At each setpoint along the trajectory, the geometric controller computes the control command, which is applied at fixed time steps to propagate the UAV state. We record rotor speeds during execution and use them to compute instantaneous power and total trajectory energy.

The total predicted energy consumption is computed as:
\begin{equation}
E=\sum_{i=1}^{n}\int P_i(t)\,dt,
\label{eq:energy_integral}
\end{equation}
where $n$ is the number of rotors and $P_i(t)$ is the instantaneous power of rotor $i$. Following~\cite{bauersfeld2022range}, we model rotor power as:
\begin{equation}
\begin{aligned}
P_i(t)=&\,6.088\times10^{-3}\omega_i(t)+1.875\times10^{-8}\omega_i^3(t) \\
&+7.700\times10^{-20}\omega_i^6(t),
\end{aligned}
\label{eq:power_model}
\end{equation}
with $\omega_i(t)$ the angular velocity of rotor $i$. This yields a physically grounded estimate of the true execution cost for each candidate without requiring real-world trial flights.

Finally, we select the next exploration goal by combining the frontier and energy-awareness stages. We first retain only feasible candidates among the high-information frontier clusters. Among these, we select the candidate with minimum predicted energy as the next goal. In this way, EAAE favours frontier regions that are both informative and energy-efficient to reach. During execution, the local planner then tracks the selected global plan while reacting to local obstacles and environmental changes. This separation enables energy-informed target selection together with robust online adaptation, although it may introduce some mismatch between the estimated offline trajectory and the ultimately executed motion.

\section{Experimental Results}
To evaluate the impact of trajectory-level energy-aware frontier selection, we implement EAAE in a full exploration pipeline and benchmark it against representative frontier baselines that do not model energy during decision-making. We design experiments to directly test our core claims: (i) EAAE reduces total mission energy by selecting informative frontiers that require less energy to execute, and (ii) this energy expense does not come at the expense of exploration performance. In addition, we report module-level runtime to quantify the computational overhead introduced by offline energy evaluation within the exploration loop. 

\subsection{Experimental Setup}
\begin{figure}[ht!]
\centering
\begin{tikzpicture}
\tikzset{every node/.style={font=\footnotesize}}
\def\l{0.45\linewidth}
\tikzstyle{img} = [inner sep=0pt, outer sep=0pt]
\tikzstyle{capt} = [
    text width = 3cm,
    align=center,
    inner sep = 0pt,
    outer sep=0pt,
    anchor=north,
    yshift=-3pt
]

\node[img] (a)
    {\includegraphics[width=\l]{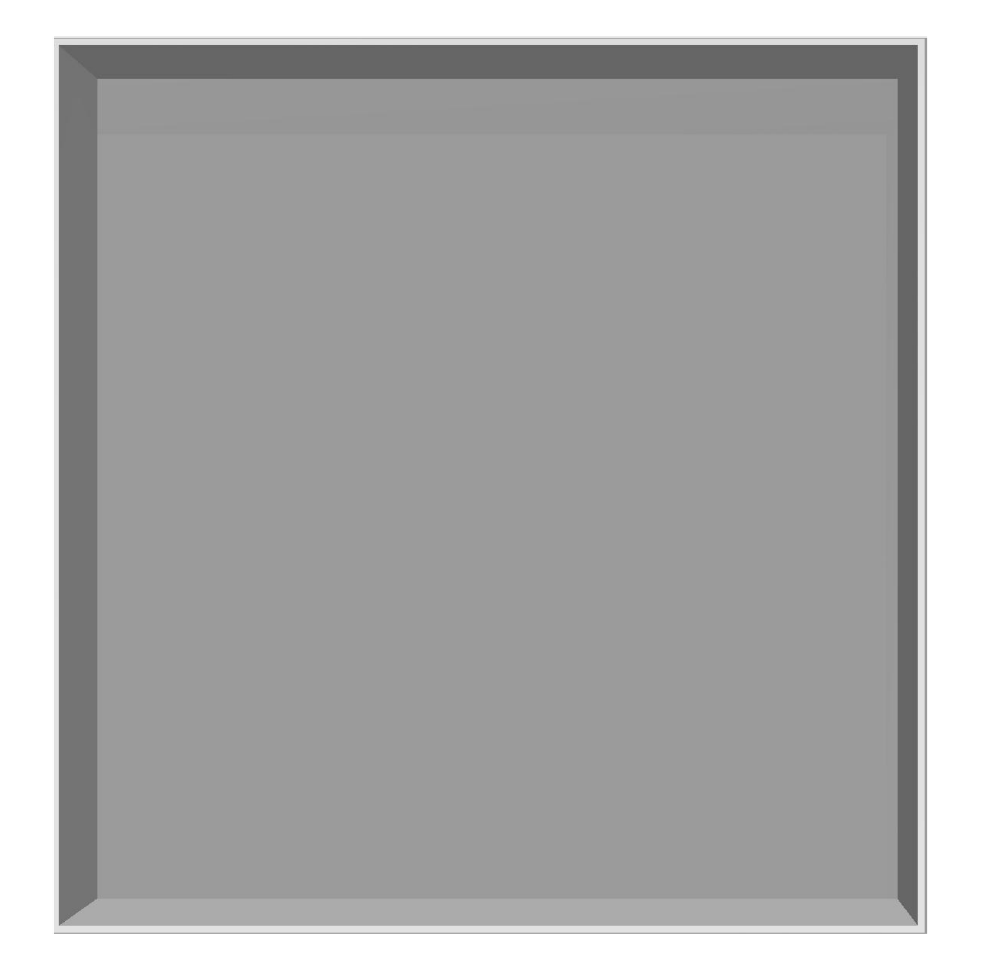}};
\node[capt] at (a.south) {(a) \textit{Simple}};

\node[img, right=6mm of a] (b)
    {\includegraphics[width=\l]{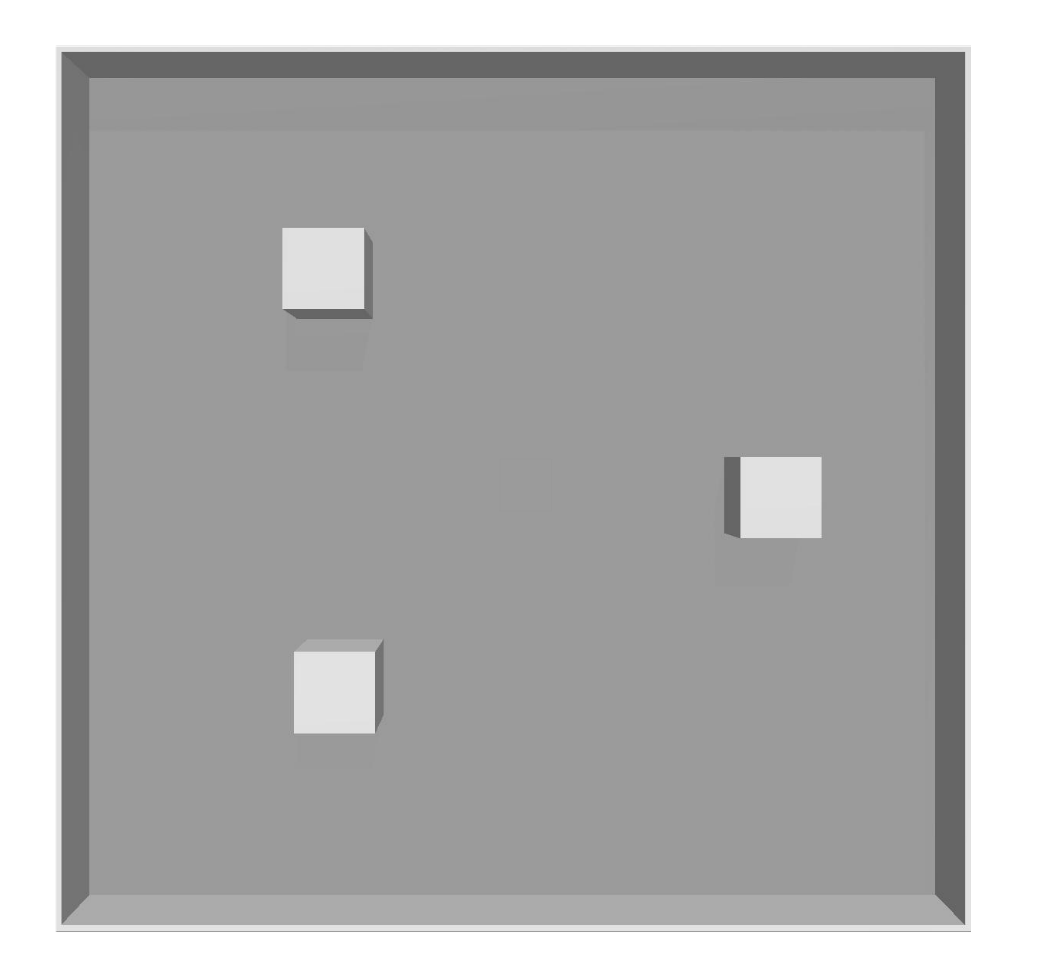}};
\node[capt] at (b.south) {(b) \textit{Pillars}};

\end{tikzpicture}
\vspace*{-6pt}
    \caption{Simulation environments used for evaluation, visualised top-down: (a) \textit{Simple}, an open room, and (b) \textit{Pillars}, a cluttered room with obstacles.}
    \label{fig:environments}
\end{figure}

We conduct our experiments in ROS Noetic with the Gazebo simulator and Agilicious~\cite{foehn2022agilicious}. Two environments are used: \textit{Simple} ($20$\,m$\times20$\,m$\times2.5$\,m) and \textit{Pillars} ($22$\,m$\times22$\,m$\times2.5$\,m), shown in Fig. \ref{fig:environments}. The UAV is equipped with a forward-facing depth sensor and and we build an online OctoMap with fixed resolution for frontier extraction and clustering.The main platform, sensing, mapping, and simulation parameters are summarised in Tab.~\ref{tab:params}.

\begin{table}[ht!]
\caption{Main simulation parameters.}
\label{tab:params}
\centering
\footnotesize
\begin{tabularx}{1\linewidth}{XX}
\toprule
Parameter & Value \\
\midrule
Rotor count / mass & $4$ / $\unit[0.752]{kg}$ \\
Max velocity / acceleration & $\unit[5.0]{m/s}$, $\unit[4.0]{m/s^2}$ \\
Depth range / horizontal FoV & $\unit[0.5-5.0]{m}$ / $\unit[1.0]{rad}$ \\
Depth frame rate & $\unit[30]{Hz}$ \\
Octomap resolution & $\unit[0.1]{m}$ \\
Control frequency & $\unit[300]{Hz}$ \\
Energy integration step & $\unit[0.02]{s}$ \\
Hardware & Intel i7-13700H, 16 GB RAM \\
\bottomrule
\end{tabularx}
\end{table}

For benchmarking, we compare EAAE against representative frontier-selection strategies while keeping the rest of the exploration pipeline identical. All methods share the same perception stack, OctoMap, frontier detection and clustering, viewpoint sampling, global trajectory generation, and local reactive execution. The only difference lies in the \emph{frontier selection rule}, ensuring a fair ablation of the proposed energy-aware decision layer.

We consider two baselines. (i) \textbf{Nearest Frontier} selects the candidate whose centroid is closest in Euclidean distance to the current UAV position, representing the standard greedy frontier heuristic~\cite{yamauchi1998frontier}. (ii) \textbf{Classic Frontier} selects the cluster with the highest frontier count, favoring large frontier regions expected to reveal more unknown space. In contrast, EAAE selects among the same feasible, high-utility candidates using the minimum \emph{predicted execution energy} obtained from offline trajectory evaluation, making the decision explicitly sensitive to the full motion profile rather than to distance or time alone.

Each method is evaluated with five independent runs per environment from the initial pose $(0,0,1)\,\mathrm{m}$. Our evaluation metrics include exploration completion time, total mission energy, map entropy (bits/cell), mean power, and module-level computation time.


\subsection{Energy Consumption and Power}

\begin{figure}[bt!]
\centering
\begin{tikzpicture}[>=stealth]
\tikzset{every node/.style={font=\footnotesize}}
\begin{axis}[
at={(0cm, 0cm)},
boxplot/draw direction = y,
title={\textbf{\textit{Simple} Environment}},
ylabel={Total Energy Consumption [kJ]},
xtick={1,2,3},
xticklabels={EAAE (ours),Nearest,Classic Frontier},
xticklabel style = {align=center, rotate=60},
width=3cm,
height=4cm,
ymajorgrids,
scale only axis,
]

\addplot+[
boxplot prepared={
lower whisker=16.5,
lower quartile=19.5,
median=21.3,
upper quartile=23.2,
upper whisker=25.5
},
blue
] coordinates {};

\addplot+[
boxplot prepared={
lower whisker=25.8,
lower quartile=27.8,
median=30,
upper quartile=32.5,
upper whisker=35
},
green!60!black
] coordinates {};

\addplot+[
boxplot prepared={
lower whisker=18.8,
lower quartile=20.5,
median=21.8,
upper quartile=23.5,
upper whisker=26.2
},
red
] coordinates {};
\end{axis}

\begin{axis}[
at={(4.5cm, 0cm)},
boxplot/draw direction = y,
title={\textbf{\textit{Pillars} Environment}},
ylabel={Total Energy Consumption [kJ]},
xtick={1,2,3},
xticklabels={EAAE (ours),Nearest,Classic Frontier},
xticklabel style = {align=center, rotate=60},
width=3cm,
height=4cm,
ymajorgrids,
scale only axis,
]
\addplot+[
boxplot prepared={
lower whisker=44,
lower quartile=44.5,
median=47.5,
upper quartile=48,
upper whisker=49.8
},
blue
] coordinates {(0,35)};

\addplot+[
boxplot prepared={
lower whisker=50,
lower quartile=50.5,
median=57,
upper quartile=63.5,
upper whisker=64
},
green!60!black
] coordinates {};

\addplot+[
boxplot prepared={
lower whisker=39.5,
lower quartile=44,
median=52.5,
upper quartile=59.5,
upper whisker=61
},
red
] coordinates {};
\end{axis}
\end{tikzpicture}
\caption{Total exploration energy over five runs per method. EAAE achieves the lowest median energy in both environments, demonstrating the benefit of trajectory-level energy-aware frontier selection.}
\label{fig:energy_boxplots}
\end{figure}
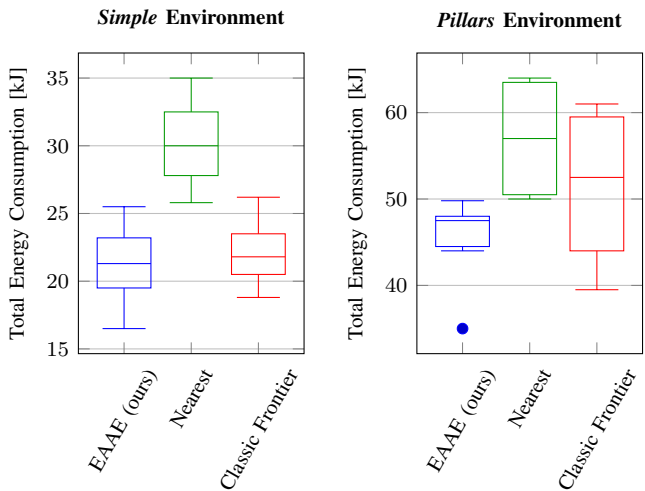

We first evaluate whether EAAE reduces the energy required to explore an unknown environment. Fig.~\ref{fig:energy_boxplots} shows the total energy consumption across repeated runs, while Tab.~\ref{tab:power} reports the corresponding average power consumption. Together, these results indicate not only how much energy each method consumes overall, but also whether the observed differences arise from longer flight duration, higher instantaneous power, or both.

\begin{table}[ht!]
\caption{Average power consumption by method.}
\label{tab:power}
\centering
\footnotesize
\begin{tabular}{lcc}
\toprule
Method & Simple (W) & Pillars (W) \\
\midrule
EAAE (ours) & 139.24 & \textbf{133.70} \\
Nearest & 167.23 & 168.36 \\
Classic Frontier & \textbf{136.33} & 134.80 \\
\bottomrule
\end{tabular}
\end{table}

EAAE is the most energy-efficient method in both environments, consuming $21.2$\,kJ in \textit{Simple} and $45.0$\,kJ in \textit{Pillars}, demonstrated in Fig. \ref{fig:energy_boxplots}. The advantage is particularly pronounced in the more cluttered \textit{Pillars} scene, where EAAE reduces total energy by $12.5$\% relative to Classic Frontier and by $21.1$\% relative to Nearest. This supports the hypothesis that trajectory-level energy-aware target selection becomes increasingly beneficial in environments where obstacle-induced detours and aggressive maneuvers make simple information gain- or geometric-based proxies less reliable.

The power measurements help explain the observed time--energy trade-off, shown in Tab. \ref{tab:power}. In particular, Nearest often completes exploration relatively quickly in cluttered scenes, but does so at substantially higher average power. By contrast, EAAE consistently maintains lower power consumption, especially in \textit{Pillars}, indicating that its selected trajectories are less demanding to execute. In \textit{Simple}, Classic Frontier achieves slightly lower average power than EAAE, but still consumes more total energy overall due to its longer completion time. This suggests that the energy savings of EAAE do not stem solely from reduced instantaneous power, but from a combination of lower-cost trajectories and competitive mission duration.

\subsection{Exploration Rate}

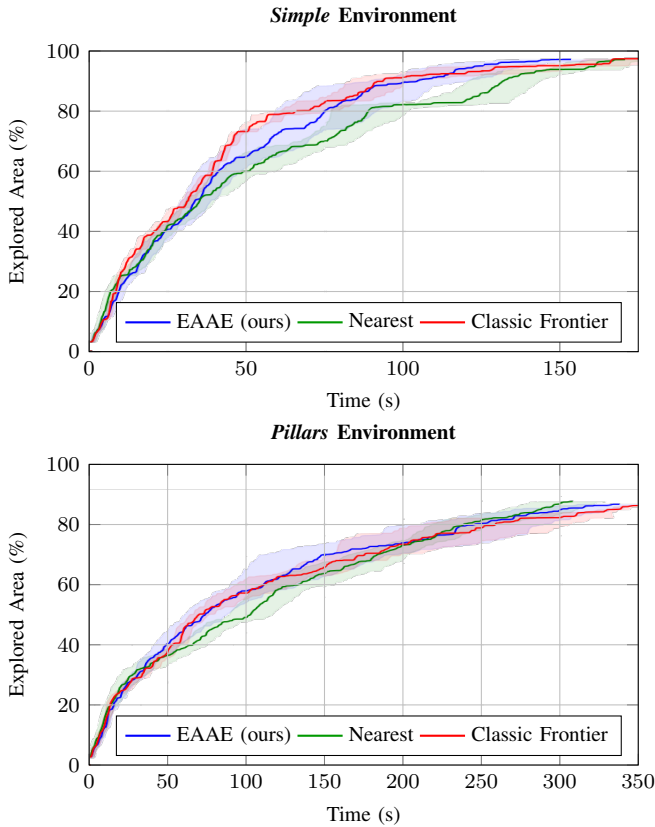
\begin{figure}[ht!]
    \centering
    \begin{tikzpicture}
    \tikzset{every node/.style={font=\footnotesize}}
    \begin{axis}[
        width=7.3cm,
        height=4cm,
        xmin=0, xmax=175,
        ymin=0, ymax=100,
        xlabel={Time (s)},
        ylabel={Explored Area (\%)},
        grid=major,
        axis on top,
        legend pos=south east,
        legend columns=3,
        scale only axis,
        title={\textbf{\textit{Simple} Environment}}
    ]
    
    \addplot [forget plot] graphics[
        xmin=0, xmax=175,
        ymin=0, ymax=100
    ] {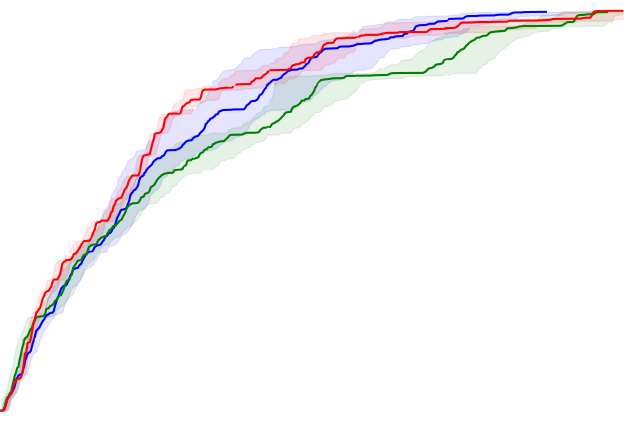};
    
    \addplot[blue, thick] coordinates {(0,0) (1,0)};
    \addlegendentry{EAAE (ours)}
    
    \addplot[green!60!black, thick] coordinates {(0,0) (1,0)};
    \addlegendentry{Nearest}
    
    \addplot[red, thick] coordinates {(0,0) (1,0)};
    \addlegendentry{Classic Frontier}
    \end{axis}
    
    \begin{axis}[
        at={(0cm, -5.5cm)},
        width=7.3cm,
        height=4cm,
        xmin=0, xmax=350,
        ymin=0, ymax=100,
        xlabel={Time (s)},
        ylabel={Explored Area (\%)},
        grid=major,
        axis on top,
        legend pos=south east,
        legend columns=3,
        scale only axis,
        title={\textbf{\textit{Pillars} Environment}}
    ]
    
    \addplot [forget plot] graphics[
        xmin=0, xmax=350,
        ymin=0, ymax=91.77,
    ] {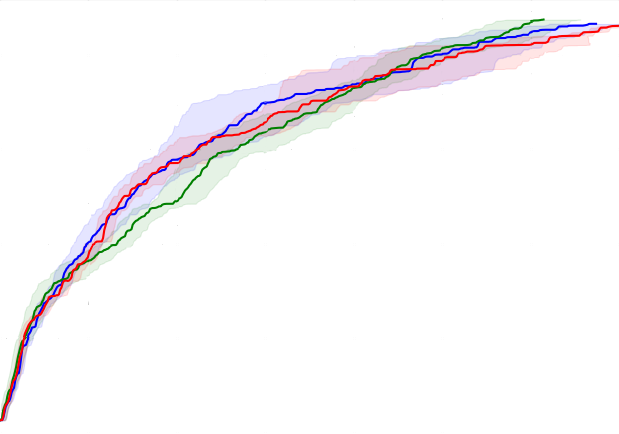};
    
    \addplot[blue, thick] coordinates {(0,0) (1,0)};
    \addlegendentry{EAAE (ours)}
    
    \addplot[green!60!black, thick] coordinates {(0,0) (1,0)};
    \addlegendentry{Nearest}
    
    \addplot[red, thick] coordinates {(0,0) (1,0)};
    \addlegendentry{Classic Frontier}
    
    \end{axis}
    \end{tikzpicture}
    \caption{Exploration progress over time. Solid lines show the means over five runs. The shaded area indicates the $\pm1\sigma$ range. EAAE maintains competitive exploration rate while reducing total energy.}
    \label{fig:exploration_rate}
\end{figure}

We next evaluate whether energy-aware frontier selection affects exploration efficiency. Fig.~\ref{fig:exploration_rate} reports the explored area percentage over time for all methods, measuring how quickly each policy converts unknown space into explored space.

In \textit{Simple}, EAAE achieves the fastest completion time ($160.0$\,s), outperforming Classic ($178.3$\,s) and Nearest (182.8~s). In \textit{Pillars}, Nearest completes fastest ($321.7$\,s), while EAAE remains competitive ($353.0$\,s) and improves over Classic ($366.6$\,s). Overall, these results show that introducing trajectory-level energy prediction into the selection loop does not slow down exploration and can match or exceed proxy-based baselines in coverage rate.

\subsection{Map Entropy and Computational Efficiency}

Finally, we show that introducing energy awareness into the exploration loop does not degrade map quality. Tab.~\ref{tab:performance} reports map entropy, where lower values indicate higher mapping certainty. In \textit{Simple}, Classic Frontier achieves the lowest entropy ($0.719$\,bits/cell), while EAAE remains close ($0.725$\,bits/cell). In \textit{Pillars}, EAAE achieves the lowest entropy ($0.754$\,bits/cell), outperforming both Nearest and Classic Frontier. Overall, EAAE maintains competitive mapping certainty and can improve map confidence in cluttered environments.

We further quantify the computational overhead of the proposed energy-aware decision layer. Tab.~\ref{tab:computation_time} summarises the runtime per exploration cycle. Global candidate trajectory generation dominates runtime, requiring $3331.6$\,ms in \textit{Simple} and 1887.0~ms in \textit{Pillars}. In contrast, frontier clustering and offline energy estimation remain lightweight, on the order of tens of milliseconds. This indicates that the main bottleneck is generating dynamically feasible trajectories for multiple candidates, while the additional cost of energy evaluation itself is comparatively small.

\begin{table}[b!]
\caption{Average computation time per EAAE module (ms).}
\label{tab:computation_time}
\centering
\footnotesize
\begin{tabular}{lccc}
\toprule
Scene & Clustering & Trajectory Gen. & Energy Est. \\
\midrule
\textit{Simple} & 17.8 & 3331.6 & 45.2 \\
\textit{Pillars} & 10.0 & 1887.0 & 34.4 \\
\bottomrule
\end{tabular}
\end{table}

Overall, these findings complement the earlier results on energy consumption and exploration progress. EAAE achieves the lowest total energy, remains competitive in exploration time, and preserves map confidence. The additional overhead of energy evaluation is modest relative to the candidate trajectory generation already required for global planning. This supports the practical feasibility of EAAE and points to trajectory generation as the main target for future runtime optimisation.

\begin{table}[ht!]
\caption{Performance statistics (mean over five runs).}
\label{tab:performance}
\centering
\footnotesize
\setlength{\tabcolsep}{4pt}
\begin{tabular}{llccc}
\toprule
Scene & Method & Explore Time (s) & Energy (kJ) & Entropy \\
\midrule
\multirow{3}{*}{\textit{Simple}}
& EAAE (ours)      & \textbf{160.0} & \textbf{21.2} & 0.725 \\
& Nearest          & 182.8          & 30.3          & 0.735 \\
& Classic Frontier & 178.3          & 22.2          & \textbf{0.719} \\
\midrule
\multirow{3}{*}{\textit{Pillars}}
& EAAE (ours)      & 353.0          & \textbf{45.0} & \textbf{0.754} \\
& Nearest          & \textbf{321.7} & 57.0          & 0.784 \\
& Classic Frontier & 366.6          & 51.4          & 0.767 \\
\bottomrule
\end{tabular}
\end{table}

\section{Conclusions and Future Work}
This paper presented Energy-Aware Autonomous Exploration (EAAE), a modular frontier-based framework that makes energy an explicit decision variable during autonomous 3D exploration with battery-limited multirotor UAVs. Our approach augments a standard exploration pipeline with a lightweight decision layer that (i) clusters frontiers into view-consistent regions, (ii) generates dynamically feasible candidate trajectories, and (iii) predicts their execution energy using an offline power-estimation loop to guide target selection. This design preserves the practicality of frontier exploration while directly accounting for the cost of the executed motion. We implemented EAAE in a complete exploration stack and evaluated it using a rotor-speed-based power model across 3D environments of increasing complexity. Results show that EAAE consistently reduces total energy consumption compared to representative distance- and information gain-based frontier baselines, while maintaining competitive exploration time and comparable map quality. Future work will focus on validating EAAE in real-world flight experiments, including in larger and dynamic environments, and extending the framework to handle wind disturbances, dynamic constraints under payload changes, and multi-UAV exploration.

\balance
\bibliographystyle{IEEEtran}
\bibliography{references}

\end{document}